\documentclass{article}

\usepackage{PRIMEarxiv}

\usepackage[utf8]{inputenc} 
\usepackage[T1]{fontenc}    
\usepackage{hyperref}       
\usepackage{url}            
\usepackage{booktabs}       
\usepackage{amsfonts}       
\usepackage{nicefrac}       
\usepackage{microtype}      
\usepackage{lipsum}
\usepackage{graphicx}
\graphicspath{{media/}}     

\usepackage{longtable}

\title{Guidelines for External Disturbance Factors in the Use of OCR in Real-World Environments  
}

\author{
  Kenji Iwata \\
  National Institute of Advanced Industrial \\Science and Technology (AIST) \\
   \And
  Eiki Ishidera \\
  NEC Corporation\\
   \And
  Toshifumi Yamaai \\
  Ricoh Co., Ltd.\\
   \And
  Yutaka Satoh \\
  National Institute of Advanced Industrial \\Science and Technology (AIST) \\
   \And
  Hiroshi Tanaka \\
  Fujitsu Limited\\
   \And
  Katsuhiko Takahashi \\
  NEC Corporation\\
   \And
  Akio Furuhata \\
  Toshiba Digital Solutions Corporation\\
   \And
  Yoshihisa Tanabe \\
   \And
  Hiroshi Matsumura\\
}

\begin{document}
\maketitle

\begin{abstract}
The performance of OCR  has improved with the evolution of AI  technology. As OCR continues to broaden its range of applications, the increased likelihood of interference introduced by various usage environments can prevent it from achieving its inherent performance. This results in reduced recognition accuracy under certain conditions, and makes the quality control of recognition devices more challenging. Therefore, to ensure that users can properly utilize OCR, we compiled the real-world external disturbance factors that cause performance degradation, along with the resulting image degradation phenomena, into an external disturbance factor table and, by also indicating how to make use of it, organized them into guidelines.
\end{abstract}


\section{Introduction}
In recent years, with the rapid and seemingly discontinuous evolution of AI  technology, recognition-based input methods, such as OCR  have experienced remarkable performance improvements year by year \cite{Xiao2023}. With the widespread adoption of smartphones and high-speed communication networks, OCR has become accessible for various users and applications, such as services that convert the content of receipts or handwritten notes into digital data by imaging them with a smartphone.

While the range of applications for OCR is expanding, the characteristic of OCR, whereby its recognition accuracy fluctuates under the influence of external disturbances in the usage environment, combined with the increased likelihood of such external disturbances as its usage broadens, has raised the risk that it may not perform well under certain conditions, resulting in degraded recognition accuracy. This situation makes quality control of recognition devices more difficult.

Therefore, to enable users to properly utilize recognition devices, the JEITA Technical Standardization Committee on Pattern Recognition, to which we belong, organized the real-world external disturbance factors that cause performance degradation in recognition-based input methods, along with the resulting phenomena of image degradation, into an external disturbance factor table and compiled guidelines by indicating how to utilize it \cite{jeita2019,jeita2023}.

\section{Guidelines for OCR Usage under Non-Uniform Illumination}

\begin{table}
 \caption{Definitions of Terms}
  \centering
  \begin{tabular}{p{4cm}p{10cm}}
    \hline
    \toprule
    \textbf{Terms}     & \textbf{Definitions}     \\
    \toprule
    Real-world environment &
    This is not just a specially prepared imaging environment suitable for OCR, but the environment in the real world. For example, environments where sunlight or indoor illumination is present, and where illumination conditions change depending on the time of day or location.     \\
    \midrule
    Camera-equipped device &
    An electronic device equipped with a camera. For example, information devices, such as tablet devices and smartphones, flatbed scanners, and document cameras.     \\
    \midrule
    Documents and forms image &
    An image of documents and forms captured with a digital camera or camera-equipped device.  \\
    \midrule
    Illumination &
    This includes not only illumination from artificial light sources but also sunlight and reflected light. \\
    \midrule
    Pixel &
    The smallest unit that makes up a digital image. \\
    \midrule
    Pixel value &
    The value that represents the luminance of a pixel. In an 8-bit grayscale image, the range is usually from 0 (black) to 255 (white). \\
    \midrule
    Blocked-up shadows &
    The loss of tone wedge in the dark areas of an image, resulting in them appearing black. \\
    \midrule
    Blown-out highlights &
    The loss of tone wedge in the bright areas of an image, resulting in them appearing white. \\
    \midrule
    Character pixels &
    Pixels that forms the character (Figure \ref{fig:fig1}). \\
    \midrule
    Background pixels &
    Pixels that forms the background (Figure \ref{fig:fig1}). \\
    \midrule
    One-character region &
    The region circumscribing the character picture elements (Figure \ref{fig:fig2}). \\
    \midrule
    Histogram &
    A histogram is obtained by dividing the range of possible pixel values into predetermined multiple intervals and
    counting the number of picture elements corresponding to each interval. \\
    \midrule
    Grade &
    A grade indicating resistance to non-uniform illumination. \\ 
    \bottomrule
  \end{tabular}
  \label{tab:table1}
\end{table}

Non-uniform illumination significantly affects OCR recognition accuracy; therefore, we first focused on this factor and established usage guidelines for OCR \cite{jeita2019}. The definitions of the terms used hereafter are shown in Table \ref{tab:table1}.

In conventional OCR using scanner input, images of uniform brightness can be obtained by blocking ambient light, installing an illumination source, and irradiating documents or forms with a fixed amount of light. In contrast, when capturing images with a digital camera or a camera-equipped device, the amount of illumination on the document or form surface may not be spatially uniform. For example, illumination light can be obstructed by the photographer’s body or items, such as the camera, causing shadows on the document or form surface, or, for illumination light with directivity, such as a spotlight light, the area around the center may be illuminated especially brightly. In images shot under such conditions, the pixel values of the picture element corresponding to the document or form surface, which should originally have uniform color and power reflection coefficient, vary significantly. Due to the characteristics of human vision, users often fail to notice these variations in pixel values; however, they can significantly affect OCR recognition accuracy.

In this guideline, non-uniform illumination is defined based on such non-uniformities in pixel values across the document or form image. Specifically, as shown in Table \ref{tab:table2}, the level of non-uniform illumination is defined in three stages, based on the distribution of pixel values for the character picture element and background picture element in the histogram of the document or form image. However, while the shape of the non-uniform illumination may also affect recognition accuracy, this influence is strongly related to each OCR algorithm and the glyph shapes of the recognized characters; therefore, it cannot be discussed in isolation and is thereby beyond the scope of this study.

\begin{figure}
    \centering
    \includegraphics[scale=1.0]{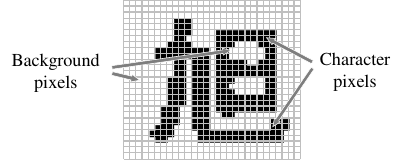}
    \caption{Examples of Character and Background Pixels}
    \label{fig:fig1}
\end{figure}

\begin{figure}
    \centering
    \includegraphics[scale=1.5]{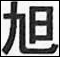}
    \caption{Example of a One-Character Region}
    \label{fig:fig2}
\end{figure}

\begin{table}
 \caption{Definition of Non-uniform Illumination Levels}
  \centering
  \begin{tabular}{p{2cm}p{12cm}}
    \hline
    \toprule
    \textbf{Level}     & \textbf{Definitions}     \\
    \toprule
    \textbf{I} &
    The distribution of pixel values for background pixels and character pixels in the histogram over the entire document or form image is separated (Figures \ref{fig:fig3} and \ref{fig:fig4}).    \\
    \midrule
    \textbf{II} &
    The pixel value distributions of background pixels and character pixels in the histogram across the entire document or form image overlap (Figures \ref{fig:fig5} and \ref{fig:fig6}). Furthermore, for every character, the pixel value distributions of background pixels and character pixels within the one-character region are separated (Figures \ref{fig:fig7} and \ref{fig:fig8}).     \\
    \midrule
    \textbf{III} &
    At least one region exists in the one-character region image where the pixel value distributions of background pixels and character pixels are not separated. Blocked-up shadows and blown-out highlights also fall under this quality category (Figures \ref{fig:fig9}, \ref{fig:fig10}, and \ref{fig:fig11}). \\
    \bottomrule
  \end{tabular}
  \label{tab:table2}
\end{table}

\begin{figure}
  \centering
  \includegraphics[scale=1.0]{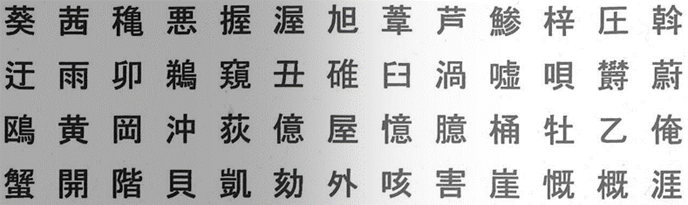}
  \caption{Example of a Document or Form Image with Level I Non-uniform Illumination}
  \label{fig:fig3}
\end{figure}

\begin{figure}
  \centering
  \includegraphics[scale=1.5]{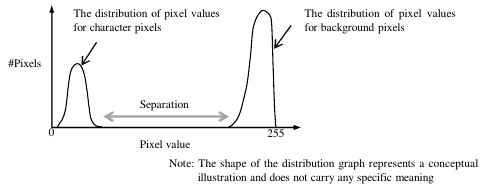}
  \caption{Example of a Histogram for a Document or Form Image with Level I Non-uniform Illumination}
  \label{fig:fig4}
\end{figure}

\begin{figure}
  \centering
  \includegraphics[scale=1.0]{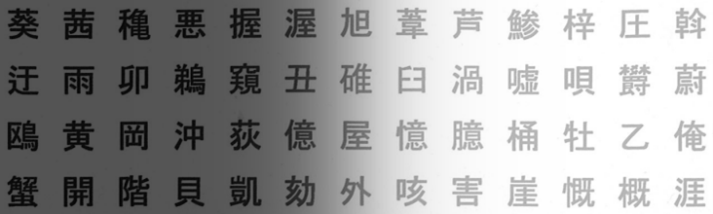}
  \caption{Example of a Document or Form Image with Level II Non-uniform Illumination}
  \label{fig:fig5}
\end{figure}

\begin{figure}
  \centering
  \includegraphics[scale=1.5]{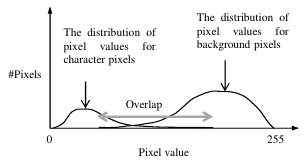}
  \caption{Example of a Histogram for a Document or Form Image with Level II Non-uniform Illumination}
  \label{fig:fig6}
\end{figure}

\begin{figure}
  \centering
  \includegraphics[scale=1.5]{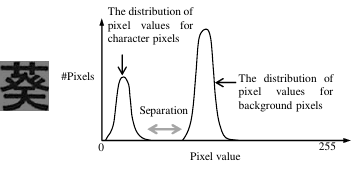}
  \caption{Example of a Histogram for a One-Character Region in a Document or Form Image with Level II Non-uniform Illumination (Low-luminance Background)}
  \label{fig:fig7}
\end{figure}

\begin{figure}
  \centering
  \includegraphics[scale=1.5]{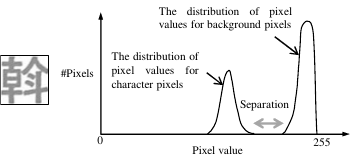}
  \caption{ Example of a Histogram for a One-Character Region in a Document or Form Image with Level II Illumination Non-Uniformity (High-luminance Background)}
  \label{fig:fig8}
\end{figure}

\begin{figure}
  \centering
  \includegraphics[scale=1.0]{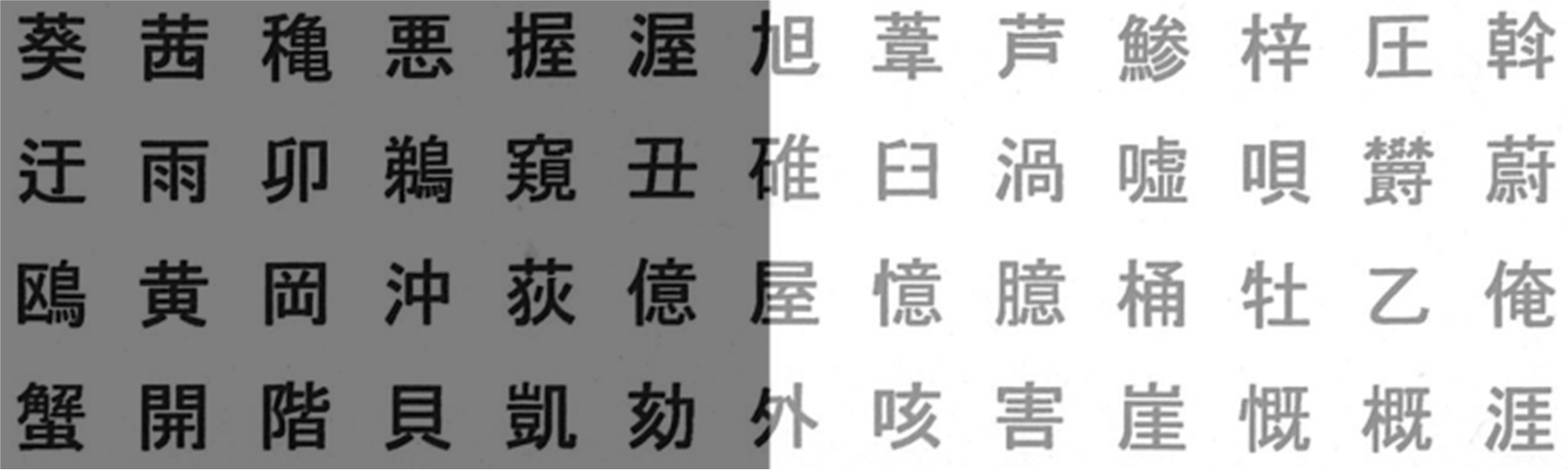}
  \caption{Example of a Document or Form Image with Non-Uniform Level III Illumination }
  \label{fig:fig9}
\end{figure}

\begin{figure}
  \centering
  \includegraphics[scale=1.5]{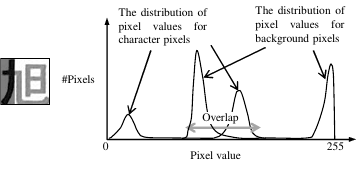}
  \caption{Example of a Histogram for a One-Character Region in a Document or Form Image with Non-Uniform Level III Illumination}
  \label{fig:fig10}
\end{figure}

\begin{figure}
  \centering
  \includegraphics[scale=1.0]{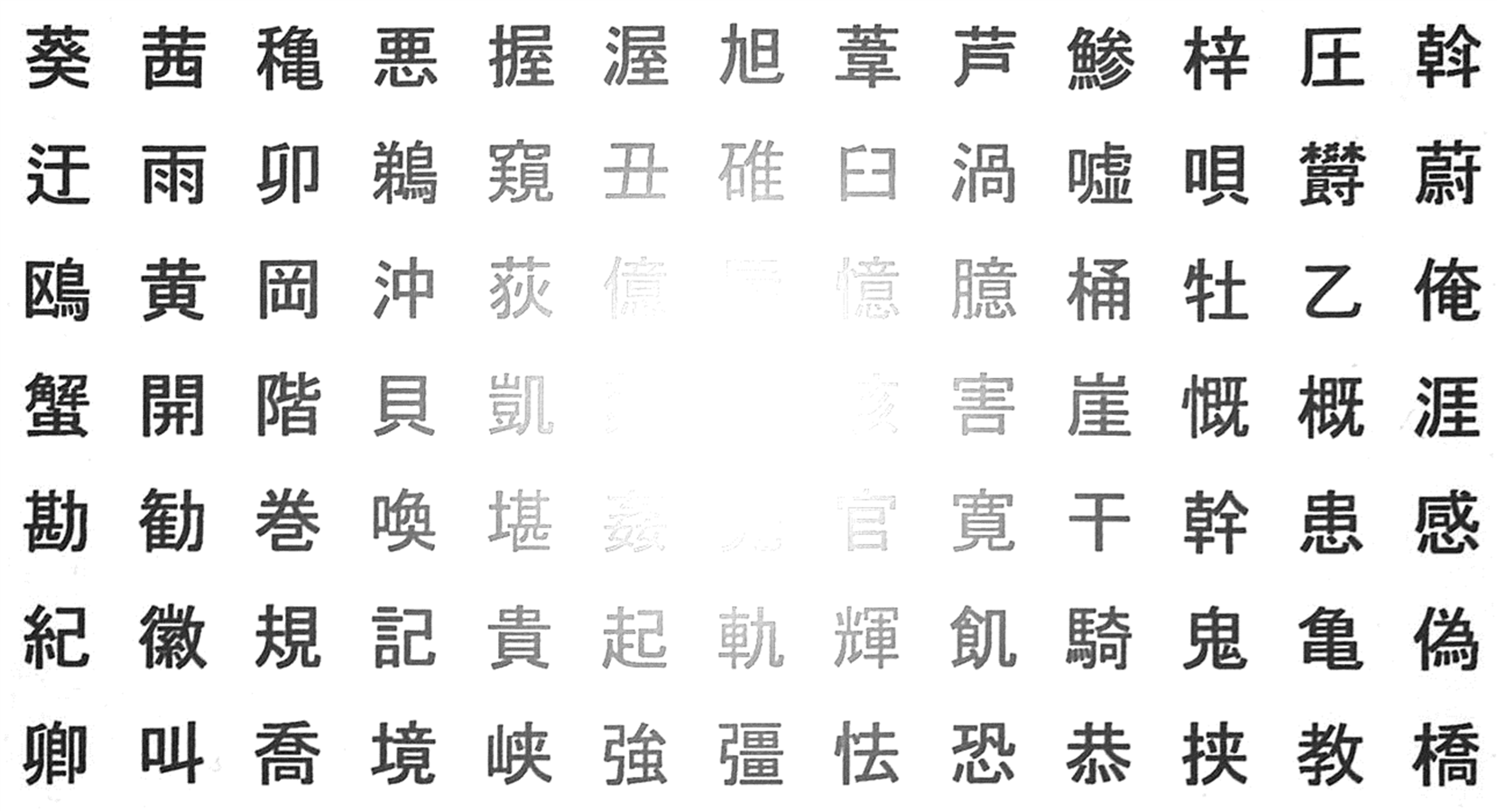}
  \caption{Example of a Document or Form Image Containing Blown Out Highlights}
  \label{fig:fig11}
\end{figure}
 
\begin{table}[h]
 \caption{Definition of OCR Grades}
  \centering
  \begin{tabular}{p{2cm}p{12cm}}
    \hline
    \toprule
    \textbf{Grade}     & \textbf{Definitions}     \\
    \toprule
    \textbf{A} &
   Functions well even in the presence of Level I illumination non-uniformity. \\
    \midrule
    \textbf{AA} &
    Functions well even in the presence of Level II illumination non-uniformity.     \\
    \midrule
    \textbf{X} &
    Functions well even in the presence of Level III illumination non-uniformity. \\
    \bottomrule
  \end{tabular}
  \label{tab:table3}
\end{table}

The OCR grades are defined in Table \ref{tab:table3} based on the levels of non-uniform illumination in Table \ref{tab:table2}. The following sections provide usage guidelines of OCR for each grade.

\begin{description}
\item[(1) Grade A OCR]~
\begin{itemize}
\item It is primarily used indoors.
\item A condition with uniform illumination that minimizes non-uniform illumination is ideal. In an environment with only one light source, keep the document/form at a distance from the light source.
\item Ensure that shadows from surrounding objects or the photographer do not overlap with any part of the document/form as much as possible. If certain parts of the document/form inevitably end up in shadow, take measures to diminish those shadows; for instance, by using multiple light sources from different positions.
 \end{itemize}
 
\item[(2) Grade AA OCR]~
\begin{itemize}
\item It can be used indoors with almost no restrictions. However, ensure not to expose it to illumination light with strong directivity, such as intense spotlight lighting. Additionally, ensure that the strong shadows created by the illumination light do not partially overlap the document or form surface.
\item When using it outdoors, take care to avoid the effects of direct sunlight. For example, ensure that the strong shadows cast by direct sunlight do not partially overlap the document or form surface.
\end{itemize}
 
\item[(3) Grade X OCR]~
\begin{itemize}
\item It can be used both indoors and outdoors with almost no restrictions.
\item However, at present, it is difficult to develop an OCR system that can operate in a general-purpose manner regardless of the occurrence of any of the three levels of non-uniform illumination. In addition, technological innovations, such as advances in camera technology and the interpolation of missing information through knowledge utilization are required.
\end{itemize}
 
\end{description}

\section{Table of External Disturbance Factors in Real-World Environments}

\begin{figure}[h]
  \centering
  \includegraphics[scale=0.25]{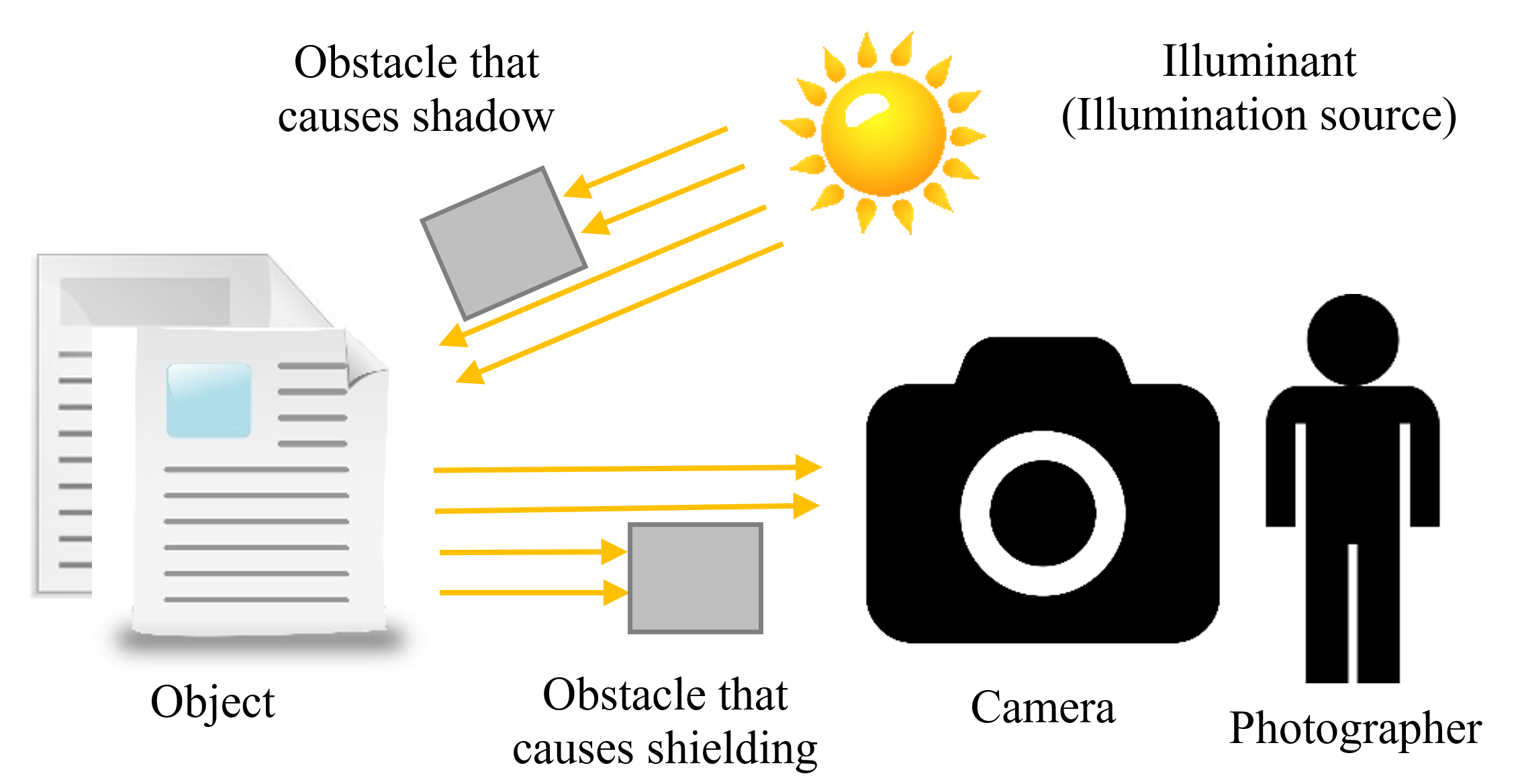}
  \caption{Process of Digitizing Document and Form Images}
  \label{fig:fig12}
\end{figure}

Figure \ref{fig:fig12} illustrates the factors that affect the image-capturing process. The light emitted from the illuminant is reflected off the object and reaches the image sensor of the camera held by the photographer. The intensity of that light is recorded as an image. In this process, obstacles may be present between the illuminant and the object or between the object and the camera. Therefore, in the digitization process of documents or form images, (a) the illuminant, (b) obstacles, (c) object, (d) camera, and (e) photographer are considered to be involved, and external disturbance factors arise from these elements. For instance, if an obstacle is present between the illuminant and the object, it creates a shadow, and if that shadow falls on a part of the object, non-uniform illumination occurs. Similarly, if an obstacle exists between the object and the camera, partial or complete shielding of the object occurs. Other types of image degradation include low contrast, low definition, fading, blocked-up shadows, defocusing, motion blur, and tilt. Various external disturbance factors related to the illuminant, object, camera, photographer, and obstacles that cause these types of image degradation were grouped by their respective sources. In addition, their relationship with the image degradation caused by the external disturbance factors was organized. Through hearings with business divisions and others that employ OCR, researchers and practitioners of OCR and image recognition discussed and compiled the findings into an external disturbance factor table [3]. Table \ref{tab:table4} presents the external disturbance factor table. Here, each external disturbance factor is assigned a code, such as L-01 or T-07, indicating its relationship to the phenomena that manifest as image degradation.

\newpage

{
\setlength{\leftmargini}{3mm} 
    
\begin{longtable}[p]{|p{25mm}|p{45mm}|p{80mm}|}
\caption{Table of External Disturbance Factors} \label{tab:table4} \\ 
\hline
    \vspace{0.5mm} \textbf{Classification} &
    \vspace{0.5mm} \textbf{Phenomenon} &
    \vspace{0.5mm} \textbf{Factor} \\
\hline
\hline
\endfirsthead
\hline
    \vspace{0.5mm} \textbf{Classification} &
    \vspace{0.5mm} \textbf{Phenomenon} &
    \vspace{0.5mm} \textbf{Factor} \\
\hline
\hline
\endhead
\hline
\endfoot
\hline
\endlastfoot
\hline
    \begin{description}
    \item \textbf{Illuminant}
    \end{description}
    &
    \begin{description}
        \item[$\bullet$] \textbf{Blocked-up shadows}
        \item L-01, L-04, L-06, L-07, L-10
        \item \includegraphics[width=3.0cm]{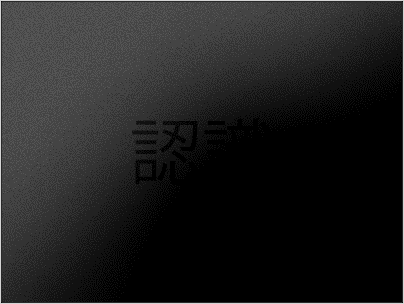}
        \item[$\bullet$] \textbf{Blown-out highlights}
        \item L-01, L-05, L-07, L-10
        \item \includegraphics[width=3.0cm]{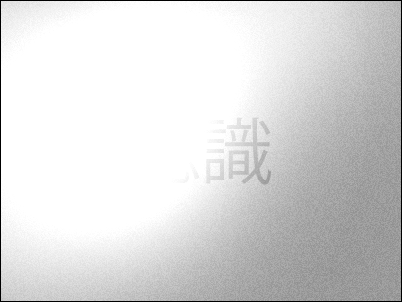}
        \item[$\bullet$] \textbf{Shading} 
        \item L-02, L-03, L-05, L-07,L-09, \item L-10
        \item \includegraphics[width=3.0cm]{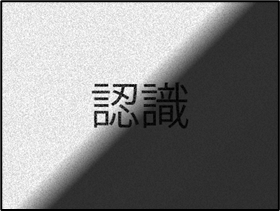}
        \item[$\bullet$] \textbf{Low contrast} 
        \item L-01, L-06, L-07, L-10
        \item \includegraphics[width=3.0cm]{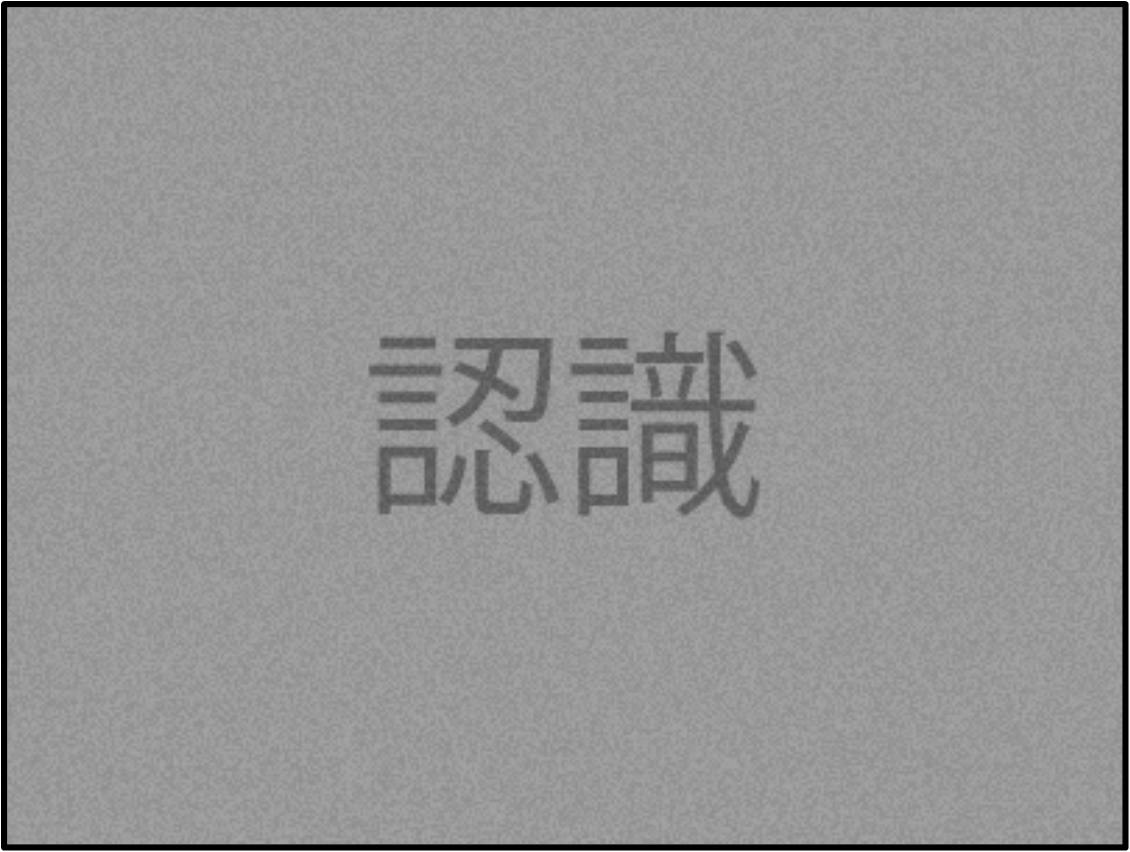}
        \item[$\bullet$] \textbf{Striped pattern} 
        \item L-04, L-10
        \item \includegraphics[width=3.0cm]{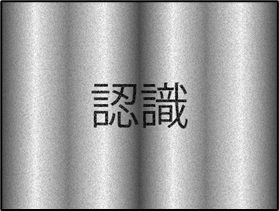}
    \end{description}
    &
    \begin{itemize}
        \item Condition and Type
        \begin{itemize}
            \item  Too bright / Too dark (L-01)
            \item  Non-uniform illumination (L-02)
            \item  Point illuminant (L-03)
            \item Frequency (Flicker) (L-04)
            \item  Flash (L-05)
            \item  Colored light (L-06)
        \end{itemize}
        \item Position and Number
        \begin{itemize}
            \item  The distance from the illuminant to the object is short (L-07)
            \item  The position is localized (L-09)
        \end{itemize}
        \item Improper use of lighting accessories (reflector, louver, diffuser, etc.) (L-10)
    \end{itemize}
    \\
    \hline

    \begin{description}
        \item \textbf{Obstacle}
    \end{description}
    &
    \begin{description}
        \item[$\bullet$] \textbf{Image distortion}
        \item O-03, O-06, O-08
        \item \includegraphics[width=3.0cm]{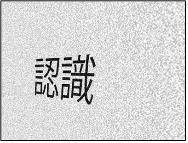}
        \item[$\bullet$] \textbf{Shadows appear}
        \item O-01
        \item \includegraphics[width=3.0cm]{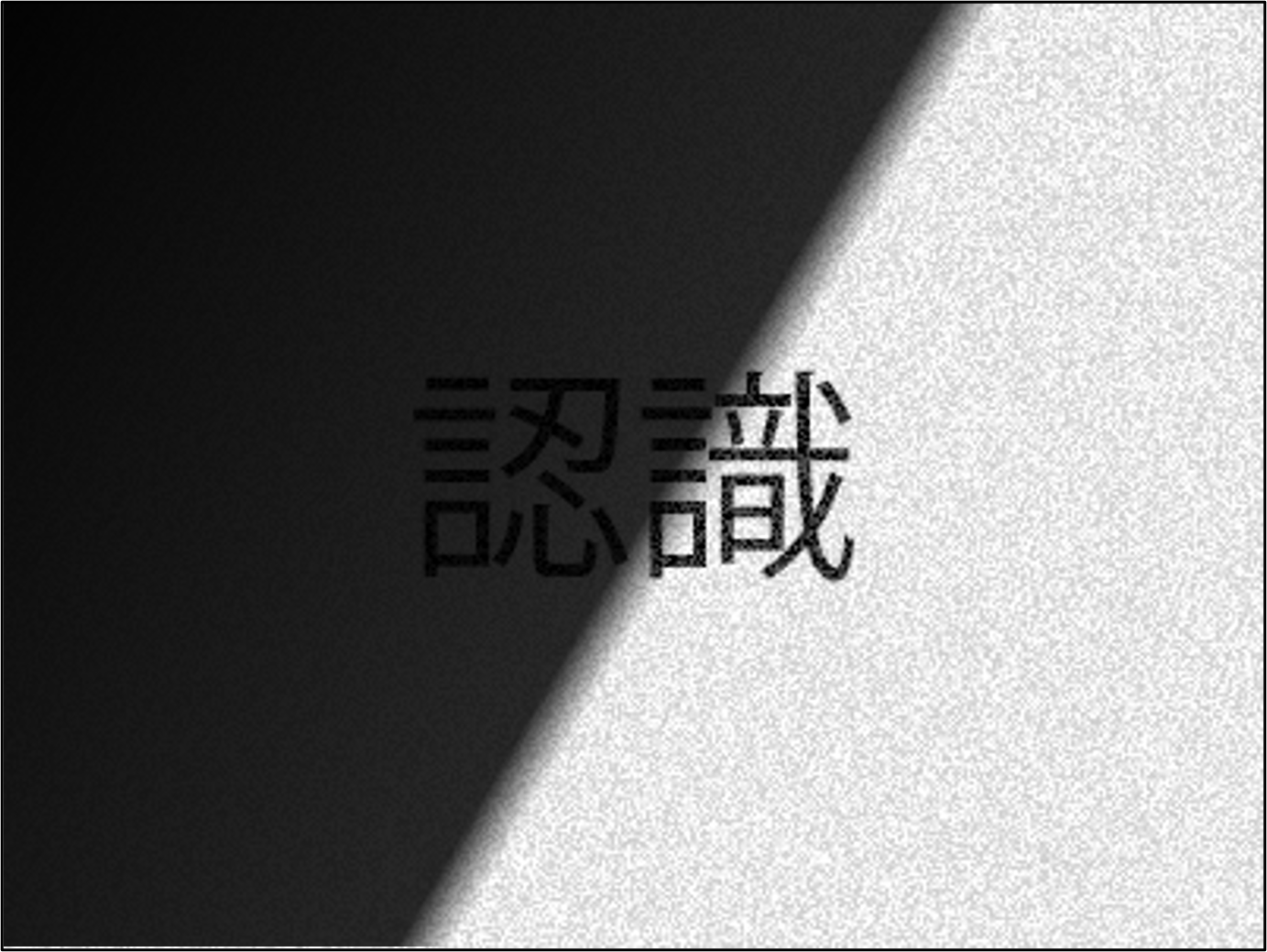}
        \item[$\bullet$] \textbf{Shielded}
        \item O-02, O-06
        \item \includegraphics[width=3.0cm]{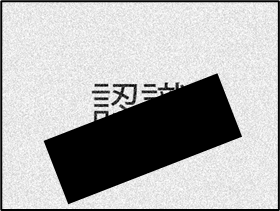}
        \item[$\bullet$] \textbf{Blurred}
        \item O-06, O-07
        \item \includegraphics[width=3.0cm]{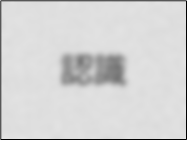}
        \item[$\bullet$] \textbf{Low contrast}
        \item O-06
        \item \includegraphics[width=3.0cm]{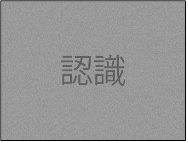}
    \end{description}
    &

    \begin{itemize}
    \item Obstacles that cause shadows (the photographer's head / hand, camera, etc.) (O-01)
    \item Obstacles that cause shielding (fingers, overlapping paper, the head of a person standing in front, etc.) (O-02)
    \item Medium (water, glass, etc.) (O-03)
    \item Obstacles in the medium (fog, rain, snow, etc.) (O-06)
    \item Foreign matter adhering to the lens (skin oil, condensation, coating, etc.) (O-07)
    \item Foreign matter adhering to the object (condensation, etc.) (O-08)
    \end{itemize}
    \\ 
    
    \begin{description}
        \item \textbf{Object}
    \end{description}
    &
    \begin{description}
        \item[$\bullet$] \textbf{Dirty}
        \item T-07
        \item \includegraphics[width=3.0cm]{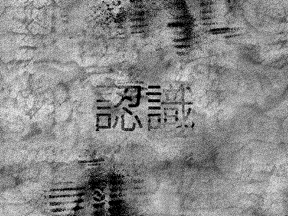}
        \item[$\bullet$] \textbf{Shiny}
        \item T-01, T-02, T-03, T-04, T-24, \item T-25
        \item \includegraphics[width=3.0cm]{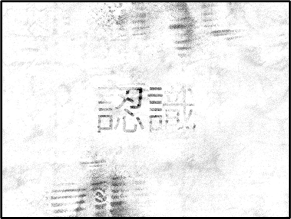}
        \item[$\bullet$] \textbf{Faint}
        \item T-13, T-16, T-17, T-25
        \item 
        \item[$\bullet$] \textbf{Distorted}
        \item T-05, T-06, T-21, T-22, T-23
        \item 
        \item[$\bullet$] \textbf{Confusing background}
        \item T-08, T-09, T-10, T-11, T-19, \item T-27, T-41, T-42, T-50, T-52, \item T-53
        \item \includegraphics[width=3.0cm]{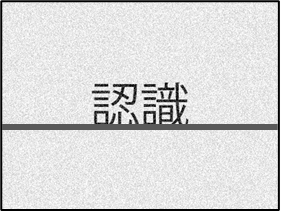}
        \item[$\bullet$] \textbf{Hard to read characters}
        \item T-19, T-23, T-25, T-31, T-34, \item T-37, T-38, T-44, T-45, T-47, \item T-51, T-54
        \item 
    \end{description}
    &
    \begin{itemize}
        \item Image/Quality Level
        \begin{itemize}
            \item Base
            \begin{itemize}
                \item Physical properties
                \begin{itemize}
                    \item Glossy paper (coated paper, carbonless copy paper, glossy paper, tracing paper, etc.) (T-01)
                    \item Monitor screen (T-02), Self-emitting light (T-03)
                    \item Projector screen (T-04)
                \end{itemize}
                \item Spatial composition
                \begin{itemize}
                    \item Three-dimensional (T-05)
                    \item Curved surface (wrinkles, waviness, folds) (T-06)
                    \item Stains and patterns
                    \item Stains (T-07)
                    \item Background pattern (copy-protection characters, patterns resembling the recognition target, background comprising multiple colors making separation from characters difficult, etc.) (T-08)
                    \item Bleed-through (T-09), Transparency (T-10)
                \end{itemize}      
                \item Scene (Background) in OCR camera
                \begin{itemize}
                    \item The recognition target is similar to the background (T-11)
                \end{itemize}
            \end{itemize}
            \item Character
            \begin{itemize}
                \item Color is not uniform (T-12)
                \item Misprinting
                \item Fading (T-13)
                \item Blocked-up shadows (T-14)
                \item Moth-eaten character image (T-15)
                \item Smudging (T-16)
                \item Ink smearing (T-17)
                \item Uneven density (T-18)
                \item Aging degradation
                \item Ink splatter missing, leaving only pressure marks (T-19)
                \item Thermal paper character degradation (T-20)
                \item Rotation (T-21)
                \item Deformation (T-22)
                \item Distorted handwritten characters, inaccurate handwritten characters (T-23)
                \item Printed characters (reflection) (T-24)
                \item Ink characteristics (faint, glossy, metallic) (T-25)
                \item Hidden characters (X-ray transmission) (T-26)
                \item Low contrast between the base and characters (T-27)
            \end{itemize}
        \end{itemize}
    \end{itemize}
    
    \\ 

    \begin{description}
        \item \textbf{Object}
    \end{description}
    &

    \begin{description}
        \item[$\bullet$] \textbf{Deteriorated character Image}
        \item T-02, T-04, T-14, T-15, T-16, \item T-17, T-18, T-20, T-49
        \item 
        \item[$\bullet$] \textbf{Difficult to separate characters}
        \item T-09, T-10, T-11, T-12, T-29, \item T-35, T-36, T-39, T-40, T-42, \item T-43, T-50
        \item \includegraphics[width=3.0cm]{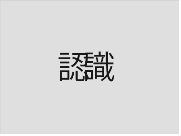}
        \item[$\bullet$] \textbf{Difficult to detect character strings}
        \item T-28, T-29, T-30, T-32, \item T-33, T-48
        \item 
        \item[$\bullet$] \textbf{Similar characters}
        \item T-46
        \item \includegraphics[width=3.0cm]{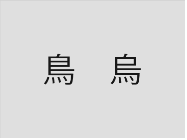}
        \item[$\bullet$] \textbf{Hidden characters}
        \item T-26
        \item 
    \end{description}
    &
    \begin{itemize}
        \item Logical level
        \begin{itemize}
            \item Layout
            \begin{itemize}
                \item Paragraph level
                \begin{itemize}
                    \item Mixed vertical and horizontal writing (T-28)
                    \item Characters present within a figure (T-29)
                \end{itemize}
                \item Character string level
                \begin{itemize}
                    \item Characters not aligned in a straight line 
                    \item Ruby character (T-30)
                    \item Superscript, Subscript (T-31)
                    \item Two-dimensionally arranged characters (chemical formulas, mathematical expressions, etc.) (T-32)
                    \item Mixed character sizes within a line (T-33)
                    \item Font variation (T-34)
                    \item Character contact (ligatures, kerning, overlap, superimposition) (T-35)
                \end{itemize}
            \end{itemize}
            \item Character
            \begin{itemize}
                \item Inverted black-and-white characters (T-36)
                \item Character type (small size) (T-37)
                \item Character decoration (embellished characters) (T-38)
                \item Overflow from entry frame (T-39)
            \end{itemize}
            \item Non-character
            \begin{itemize}
                \item Superimposition with ruling lines (T-40)
                \item Superimposition with a stamp (T-41)
                \item Shading (T-42)
                \item Underline (T-43)
            \end{itemize}
        \end{itemize}
        \item Recognition Target (Category)
        \begin{itemize}
            \item Character
            \begin{itemize}
                \item Language <Japanese entry field, English entry field, etc.> (T-44)
            \end{itemize}
            \item Character type
            \begin{itemize}
                \item Uncommon typeface (T-45)
                \item Variant characters (T-46)
                \item Special font (T-47)
                \item Special characters used in chemical formulas, mathematical expressions, etc. (T-48)
                \item Character decoration
                \item Dot characters (T-49)
                \item Engraved characters (T-50)
                \item Logo (T-51)
                \item Three-dimensional character (T-52)
                \item Embossed (T-53)
                \item Artistic characters (T-54)
            \end{itemize}
        \end{itemize}
    \end{itemize}
    \\
    
    \begin{description}
        \item \textbf{Camera /}
        \item \textbf{Photographer}
    \end{description}
    &
    \begin{description}
        \item[$\bullet$] \textbf{Blocked Up Shadows}
        \item C-03, C-04, C-06, C-10, C-13,  \item C-14, C-15, C-24
        \item \includegraphics[width=3.0cm]{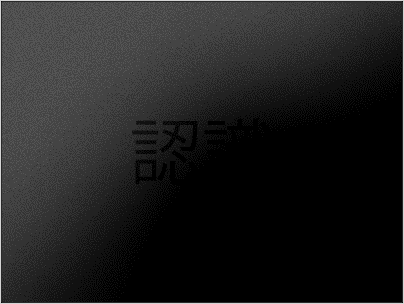}
        \item[$\bullet$] \textbf{Blown out highlights}
        \item C-03, C-10, C-11, C-13, C-14,  \item C-15, C-24
        \item \includegraphics[width=3.0cm]{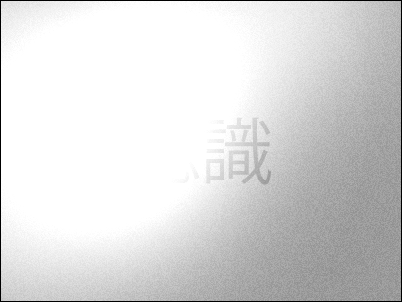}
        \item[$\bullet$] \textbf{Out of focus}
        \item C-03, C-07, C-08, C-23
        \item \includegraphics[width=3.0cm]{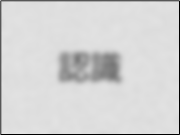}
        \item[$\bullet$] \textbf{Double (multiple) appearance}
        \item C-06, C-08, C-19, C-22
        \item \includegraphics[width=3.0cm]{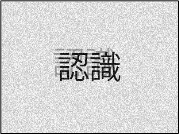}
    \end{description}
    &
    \begin{itemize}
        \item Optical system issues
        \begin{itemize}
            \item Lens characteristics
            \begin{itemize}
                \item Low definition (C-01)
                \item Too wide-angle (C-02)
                \item Lens aperture (C-03)
            \end{itemize}
            \item Peripheral image quality degradation
            \begin{itemize}
                \item Light falloff (C-04)
                \item Definition reduction (C-05)
                \item Internal reflection (C-06)
                \item Insufficient depth of field (C-07)
                \item Aberration (C-08)
            \end{itemize}
        \end{itemize}
        \item Imaging sensor
        \begin{itemize}
            \item Low picture element count (C-09)
            \item Narrow dynamic range (C-10)
            \item Smear (C-11)
            \item Rolling shutter (C-12)
        \end{itemize}
        \item Mechanical system
        \begin{itemize}
            \item Overexposure/underexposure (C-13)
            \item too high/too low Exposure Value (C-14)
            \item Too Small Aperture / Too Large Aperture (C-15)
            \item Shutter speed too fast/too slow (C-16)
        \end{itemize}
        \item Image information processing
        \begin{itemize}
            \item Coding and compression noise (C-17)
            \item Low contrast of character background due to high compression PDF (C-18)
            \item Failure of multiple exposure processing (C-19)
        \end{itemize}
        \item Shooting conditions
        \begin{itemize}
            \item Too far distance (C-20)
            \item The viewpoint is not facing the subject directly (C-21)
            \item Camera stability (camera shake) (C-22)
            \item Camera setting error
            \item Focus (C-23)
            \item Exposure (C-24)
            \item Zoom (C-25)
        \end{itemize}
    \end{itemize}

        \\
    
    \begin{description}
        \item \textbf{Camera /}
        \item \textbf{Photographer}
    \end{description}
    &
    \begin{description}
        \item[$\bullet$] \textbf{Distortion}
        \item C-02, C-10, C-12, C-21
        \item \includegraphics[width=3.0cm]{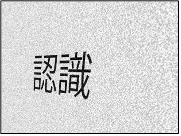}
        \item[$\bullet$] \textbf{Low contrast}
        \item C-04, C-10, C-13, C-14
        \item \includegraphics[width=3.0cm]{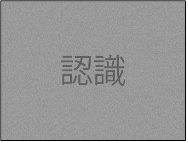}
        \item[$\bullet$] \textbf{Low definition}
        \item C-01, C-02, C-09, C-20, C-25
        \item \includegraphics[width=3.0cm]{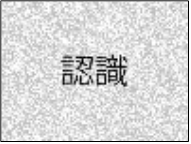}
        \item[$\bullet$] \textbf{Noise (white noise, striped pattern noise, block noise)}
        \item C-11, C-16, C-17
        \item \includegraphics[width=3.0cm]{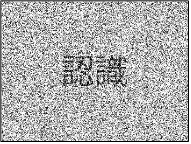}
    \end{description}
    &
    \\ 
\end{longtable}
}
\newpage

\section{Utilization of the Factor Table}
The above factor table can be applied to a wide range of uses, and the following are some possible applications.

\begin{itemize} 
\item[(a)] When OCR engineers develop OCR for a specific application, they can learn the axes of training data variations (augmentations) that need to be prepared.
\item[(b)] When a user applies OCR to a certain purpose and finds the recognition accuracy insufficient, they can identify the factors that need to be rechecked based on the condition of the captured images.
\item[(c)] When a user intends to utilize OCR for a particular application, they can discover the factors that should be considered based on the usage environment.
\end{itemize}
Here, to illustrate how to utilize the factor table, examples (a) and (b) are presented as case studies, demonstrating the specific usage methods.

\subsection{Data Augmentation}

\begin{figure}
    \centering
    \includegraphics[scale=1.4]{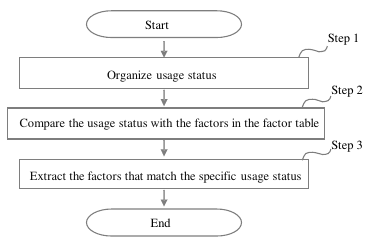}
    \caption{Example of a Flowchart for Understanding the Variations of Training Data to be Prepared}
    \label{fig:fig13}
\end{figure}

When OCR engineers develop OCR for a specific application, they can learn the axes of training data variations (augmentations) that should be prepared. For example, by following the procedure shown in Figure \ref{fig:fig13}, one can narrow down the necessary variations of training data from the factor table. Below, as an example, we assume a case of developing OCR to read prices printed on a supermarket flyer and provide a detailed explanation of each step in the process.

\textbf{Step 1: Organize the usage conditions}

First, organize the usage conditions of the OCR to be developed. For instance, to develop an OCR for reading the prices printed on a supermarket flyer, the usage conditions include the following:
\begin{itemize} 
\item Laying the flyer as flat as possible on an office desk
\item Standing by the desk while holding a smartphone to image the flyer
\end{itemize}
These conditions should be organized.

\textbf{Step 2: Compare the usage conditions to the factors in the factor table}

Compare the usage conditions organized in Step 1 with the factors in the factor table and examine whether or not they match. For example, regarding factors related to obstacles, it can be organized as shown in Table \ref{tab:table5}. Similarly, compare and organize the usage conditions with respect to the factors of the illuminant, object, and camera/photographer.

\textbf{Step 3: Extract factors that match the specific usage conditions}

Extract only the factors that match the usage conditions from the organization in Step 2. The extracted factors indicate the axes of the training data variations that should be prepared. For example, as shown in Table 5, variations should be prepared for obstacles that cause shadows and shielding.

\begin{table}
 \caption{Example of Matching the Factors in the Usage Status with the Factors in the Factor Table}
  \centering
  \begin{tabular}{p{6cm}p{1cm}p{4.5cm}p{2.4cm}}
    \toprule
    \textbf{Factor} &
    \textbf{Factor No.} &   
    \textbf{Specific usage status} &   
    \textbf{Whether the factors match the usage status}
    \\
    \hline
    \midrule
    \textbf{Obstacles that cause shadows (such as the photographer (head, hand), camera, etc.)} &
    \textbf{O-01} &
    \textbf{Possibility of shadows being cast by the photographer or similar factors} &
    \textbf{Match}
    \\
    \midrule
    \textbf{Obstacles that cause shielding (such as fingers, overlapping paper, or the head of a person standing in front)} &
    \textbf{O-02} &
    \textbf{Possibility of occlusion caused by straps or similar objects} &
    \textbf{Match}
    \\
    \midrule
    Medium (such as water, and glass.)	&
    O-03 &
    An LED illumination with almost no directivity &
    Mismatch
    \\
    \midrule
    Obstacles in the medium (such as fog, rain, and snow) &
    O-06 &
    No influence from weather because it is indoors &
    Mismatch
    \\
    \midrule
    Adhesion on the lens (such as fingerprints, condensation, and coatings)	&
    O-07 &
    The lens will be kept clean and used &
    Mismatch
    \\
    \midrule
    Adhesion on the subject (such as condensation) &
    O-08 &
    No substances adhered to the flyer &
    Mismatch
    \\
    \bottomrule
  \end{tabular}
  \label{tab:table5}
\end{table}

\subsection{Reconfirmation of Factors for OCR Users}

\begin{figure}
    \centering
    \includegraphics[scale=1.4]{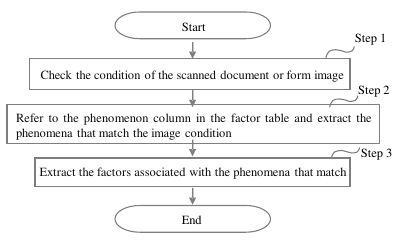}
    \caption{Example of a Flowchart for Identifying Considerations during Shooting based on the Situation of Scanned Documents and Form Images}
    \label{fig:fig14}
\end{figure}

When a user has employed OCR for a particular purpose and found that the recognition accuracy was insufficient, they can identify the factors that need to be rechecked based on the condition of the shot images. For example, by following the procedure shown in Figure \ref{fig:fig14}, one can narrow down the considerations during shooting.

\textbf{Step 1: Check the condition of the scanned document or form images}

Check the condition of the scanned document or form images (the degradation phenomena observed in the images). For example, regarding the image condition, extract items such as
\begin{itemize}
\item Blown out highlights
\item Non-uniform illumination (shading)
\item Shiny areas on the object 
\end{itemize}

\textbf{Step 2: Refer to the phenomenon column in the factor table and extract the phenomena that match the image condition}

By referring to the phenomenon column in the factor table, extract the phenomena that match the image conditions identified in Step 1. An example of the extraction results while assuming the above image conditions is shown in Table \ref{tab:table6}.

\textbf{Step 3: Extract the factors associated with the matching phenomena}

Refer to the factor table and extract the factors associated with the phenomena extracted in Step 2. For the example shown in Table \ref{tab:table6}, the factors listed in Table \ref{tab:table7} are extracted. The factors extracted here indicate the considerations during shooting. If any of these factors coincide with the current OCR usage conditions, examine whether the occurrence of those factors can be prevented. For example, if the flash was used during imaging, try imaging without using the flash.

{
\setlength{\leftmargini}{3mm} 

\begin{table}
    \caption{Example of Extracting Phenomena that Match the Image Condition}
    \centering
    \begin{tabular}{p{35mm}p{70mm}}
        \toprule
        \textbf{Classification} &
        \textbf{Phenomenon} 
        \\
        \hline
        \midrule
        Illuminant
        &
        \begin{tabular}{l}
            $\bullet$  Blown-out highlights\\
            \quad L-01, L-05, L-07, L-10\\
            $\bullet$  Shading\\
            \quad L-02, L-03, L-05, L-07, L-09, L-10\\
        \end{tabular}
        \\
        \midrule
        Obstacle 
        &
        \\
        \midrule
        Object 
        &
        \begin{tabular}{l}
            $\bullet$  Shiny\\
            \quad T-01, T-02, T-03, T-04, T-24, T-25\\
        \end{tabular}
        \\
        \midrule
        
        Camera / Photographer
        &
        \begin{tabular}{l}
            $\bullet$ Blown-out highlights\\
            \quad C-03, C-10, C-11, C-13, C-14, C-15, C-24\\
        \end{tabular}
        \\
        \bottomrule
    \end{tabular}
    \label{tab:table6}
\end{table}

\begin{table}
    \caption{Refer to the Factor Table and Extract Factors Associated with the Phenomena Extracted in Step 2}
    \centering
    \begin{tabular}{|p{20mm}|p{45mm}|p{80mm}|}
        \hline
        \vspace{0.5mm} \textbf{Classification} &
        \vspace{0.5mm} \textbf{Phenomenon} &
        \vspace{0.5mm} \textbf{Factor} 
        \\
        \hline
        \hline
        \vspace{1pt}
        \textbf{Illuminant}
        &
        \vspace{-5pt}
        \begin{description}
            \item[$\bullet$] \textbf{Blocked-up shadows}
            \item L-01, L-04, L-06, L-07, L-10
            \item \includegraphics[width=2.0cm]{s1.png}
            \item[$\bullet$] \textbf{Blown-out highlights}
            \item L-01, L-05, L-07, L-10
            \item \includegraphics[width=2.0cm]{s2.png}
            \item[$\bullet$] \textbf{Shading} 
            \item L-02, L-03, L-05, L-07,L-09, \item L-10
            \item \includegraphics[width=2.0cm]{s3.png}
        \end{description}
        \vspace{-5pt}
        &
        \vspace{-5pt}
        \begin{itemize}
            \item Condition and Type
            \begin{itemize}
                \item  Too bright / Too dark (L-01)
                \item  Non-uniform illumination (L-02)
                \item  Point illuminant (L-03)
                \item  Flash (L-05)
                \item  Colored light (L-06)
            \end{itemize}
            \item Position and Number
            \begin{itemize}
                \item  The distance from the illuminant to the object is short (L-07)
                \item  The position is localized (L-09)
            \end{itemize}
            \item Improper use of lighting accessories (reflector, louver, diffuser, etc.) (L-10)
        \end{itemize}
        \vspace{-5pt}
        \\
        \hline
        \vspace{1pt}
        \textbf{Object}
        &
        \vspace{-5pt}
        \begin{description}
            \item[$\bullet$] \textbf{Shiny}
            \item T-01, T-02, T-03, T-04, T-24, \item T-25
            \item \includegraphics[width=2.0cm]{t2.png}
        \end{description}
        \vspace{-5pt}
        &
        \vspace{-5pt}
        \begin{itemize}
            \item Base
            \begin{itemize}
                \item Physical properties
                \begin{itemize}
                    \item Glossy paper (coated paper, carbonless copy paper, glossy paper, tracing paper, etc.) (T-01)
                    \item Monitor screen (T-02), Self-emitting light (T-03)
                    \item Projector screen (T-04)
                \end{itemize}
            \end{itemize}
            \item Character
            \begin{itemize}
                \item Printed characters (reflection) (T-24)
                \item Ink characteristics (faint, glossy, metallic) (T-25)
            \end{itemize}
        \end{itemize}
        \vspace{-5pt}
        \\
        \hline
        \vspace{1pt}
        \textbf{Camera / Photographer}
        &
        \vspace{-5pt}
        \begin{description}
            \item[$\bullet$] \textbf{Blown out highlights}
            \item C-03, C-10, C-11, C-13, C-14,  \item C-15, C-24
            \item \includegraphics[width=2.0cm]{c2.png}
        \end{description}
        \vspace{-5pt}
        &
        \vspace{-5pt}
        \begin{itemize}
            \item Optical system issues
            \begin{itemize}
                \item Lens characteristics
                \begin{itemize}
                    \item Lens aperture (C-03)
                \end{itemize}
            \end{itemize}
            \item Imaging sensor
            \begin{itemize}
                \item Narrow dynamic range (C-10)
                \item Smear (C-11)
            \end{itemize}
            \item Mechanical system
            \begin{itemize}
                \item Overexposure/underexposure (C-13)
                \item too high/too low Exposure Value (C-14)
                \item Too Small Aperture / Too Large Aperture (C-15)
            \end{itemize}
            \item Shooting conditions
            \begin{itemize}
                \item Camera setting error
                \item Exposure (C-24)
            \end{itemize}
        \end{itemize}
        \vspace{-5pt}
        \\
        \hline
    \end{tabular}
    \label{tab:table7}
\end{table}
}

\section{Conclusion}
Although external disturbances can degrade the performance of recognition systems, such as OCR, no clear systematic organization has been proposed to date. Therefore, the phenomena and factors that cause performance degradation in real-world environments are comprehensively enumerated as an external disturbance factor table, and guidelines for its utilization are organized. We hope that these guidelines will contribute to the broader adoption of recognition systems, such as OCR.

\section*{Acknowledgments}
We would like to thank former Professor Kazuhiko Yamamoto of Gifu University, who established this committee by focusing on recognition-based input methods in real-world environments. We thank all the relevant parties from Hitachi, Ltd., Canon Inc., Sharp Corporation, and the Japan Newspaper Publishers \& Editors Association (KYODO NEWS) for their participation in the discussions as committee members. We appreciate former Professor Takio Kurita of Hiroshima University for his long-term participation in our discussions as a visiting member. We thank all those involved with the Japan Electronics and Information Technology Industries Association, as well as Mr. Shiokawa, Mr. Kitada, and Mr. Yoshida of the secretariat.

\bibliographystyle{unsrt}

\end{document}